\documentclass[cameraready]{Interspeech}

\newcommand{\ourmethod}{{\textsc{Progress}}}

\title{\ourmethod: Coverage-guided RL to Train Search-augmented LLM Agent}

\author[affiliation={1}, correspondingauthor]{Sudipta}{Paul}
\author[affiliation={1}]{Vijay}{Srinivasan}
\author[affiliation={1}]{Vivek}{Kulkarni}
\author[affiliation={1}]{Aounon}{Kumar}
\author[affiliation={1}]{Yashas Malur}{Saidutta}
\author[affiliation={1}]{Wenbo}{Li}
\author[affiliation={1}]{Srinivas}{Chappidi}

\address{$^1$ AI Center-Mountain View, Samsung Electronics }

\email{\{sudipta.paul, v.srinivasan, v.kulkarni1,  aounon.kumar, ym.saidutta, wenbo.li1, vasu.c\}@samsung.com}

\keywords{Search-augmented Agent, Multi-hop QA, Reinforcement Learning, Coverage Reward, Query Decomposition}

\usepackage{comment}

\newcommand{\sudipta}[1]{\textbf{\textcolor{blue}{SP: #1}}}

\usepackage{graphicx}
\usepackage{caption}
\usepackage{subcaption}
\usepackage{multirow}

\begin{document}

\maketitle

\begin{abstract}

Existing search-augmented LLM agents are trained using Reinforcement Learning to boost its reasoning capabilities. However, these approaches primarily rely on outcome-level rewards which provide little supervision over search behavior and overlook agent's ability to decompose complex queries properly. To mitigate this issue, we propose \ourmethod~which utilizes teacher-guided coverage reward to explicitly shape decomposed query generation of the policy model. During training, frozen teacher models are used to decompose complex queries into essential search queries. These essential search queries are utilized to guide the search behavior of the policy model. Integrated into an R1-style training framework, our approach provides lightweight guidance over query decomposition decisions without dense process-level supervision. Experiments show that coverage-guided RL improves overall task performance, highlighting the importance of explicitly supervising search strategies in agentic LLMs.    
\end{abstract}

\section{Introduction}

Large language models augmented with external search tools can execute complex and multi-turn reasoning tasks based on a user query by retrieving real-time, external information \cite{jin2025search, zheng2025deepresearcher}. In such agentic setting, the model must learn not only how to reason over retrieved evidence, but also what to search and when to search. Recent works \cite{jin2025search} show that training small LLMs with reinforcement learning (RL) to interleave reasoning and search actions can significantly improve performance on multi-hop question answering and knowledge-intensive tasks . However, despite encouraging empirical progress, current RL-based search agents largely rely on outcome-level rewards (e.g., exact match accuracy), providing little direct supervision over the quality of decomposed search queries. This limitation leads to inefficient search behavior, generation of composite search queries, and memorization of reasoning patterns, especially in agents with small language models.

\begin{figure}[t]
    \centering
    \includegraphics[width=\linewidth]{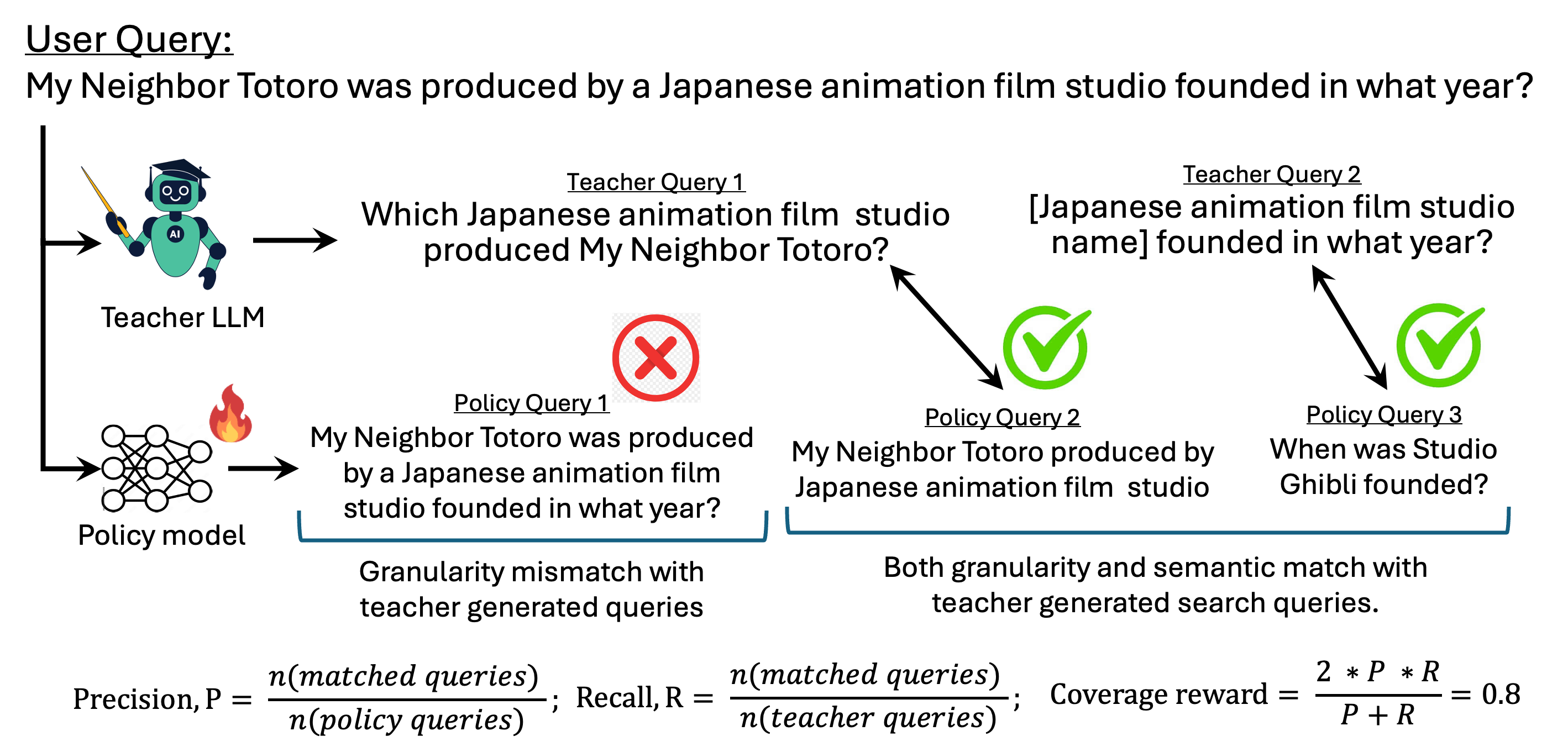}
    \caption{This figure illustrates the computation of coverage reward based on query matching information. The Policy Query 1 does not match with any Teacher Query due to granularity mismatch. However, both Policy Query 2 and 3 match with Teacher Queries in terms of semantics and granularity.}
    \vspace{-5pt}
    \label{fig:reward}
\end{figure}

To enable small language models with granular query decomposition with sufficient coverage over required reasoning hops, we propose \ourmethod.~It is a lightweight yet effective framework that utilizes teacher-generated essential search queries to guide the policy model. Our key insight is that successful multi-hop reasoning depends on successful decomposition of complex queries and comprehensive coverage over essential set of latent information needs. We also observe that strong LLM can approximate such search behavior significantly better than small language models. This observation motivates us to utilize a strong LLM as a guide for the policy model training. 

Instead of supervising each reasoning step, we first generate a set of essential search queries using a frozen teacher model. Then we identify the matches between generated search queries of policy model and pre-generated essential search queries of the teacher model based on both semantics and granularity. To encourage the policy model to generate search queries that are both comprehensive and focused, we introduce a trajectory-level \emph{coverage reward} based on the harmonic mean of precision and recall (F1-score) computed on matched queries as illustrated in Fig. \ref{fig:reward}. \ourmethod~builds upon the Search-R1 training paradigm \cite{jin2025search}, which treats the search engine as part of the environment and stabilizes RL training through retrieved-token masking and KL-regularized policy optimization. Compared to outcome-only training, coverage reward in \ourmethod~introduces an explicit inductive bias toward effective retrieval strategies without requiring dense step-level manual annotation. Empirical results suggest that explicitly shaping search behavior is a critical ingredient in training robust search-augmented LLM agents. Our contributions are summarized as follows:

\begin{itemize}

\item We introduce \ourmethod, a teacher-guided training framework that provides structured supervision of the search behavior for a RL-trained search-augmented LLM agent.

\item We propose coverage reward which is formulated based on the overlap between policy-generated search queries and teacher generated essential queries. We integrate it into the RL training framework, preserving stability while improving search quality without step-level manual supervision.

\item We empirically show that \ourmethod~leads to 2-5$\%$ absolute improvement across different multi-hop QA datasets compared to baseline while preserving/improving accuracy for general QA datasets.  

\end{itemize}

\section{Related Works}

Recent works \cite{jin2025search, zheng2025deepresearcher} have explored using large language models (LLMs) as search-augmented agents. It has enabled open-domain question answering \cite{chen2017reading} by LLMs through the retrieval of evidence from large corpora and the generation of answers based on the retrieved context. Based on the complexity of the user query, open-domain question answering can be categorized into two sub-classes: general question answering \cite{kwiatkowski2019natural} (focuses on answering from a single source of information) and multi-hop question answering \cite{yang2018hotpotqa} (requires linking multiple, distinct pieces of information—often across different sources). To address open-domain question answering, several approaches interleave reasoning steps with tool calls at inference time. IRCoT \cite{trivedi2023interleaving} explicitly interleaves retrieval with chain-of-thought. ReAct \cite{yao2022react} prompts LLMs to alternate between reasoning traces and actions (e.g., Wikipedia search). Few works \cite{wang2025chain, guan2025deeprag, asai2023self} train agentic RAG models using supervised fine-tuning (SFT) on multi-step retrieval trajectories synthesized by stronger models or search procedures. Chain-of-Retrieval Augmented Generation (CoRAG) \cite{wang2025chain} constructs intermediate retrieval chains via rejection sampling. DeepRAG \cite{guan2025deeprag} models retrieval-augmented reasoning as an MDP and uses search-based data generation (including tree search) to produce training trajectories. Recently, Reinforcement Learning has been used to optimize end-to-end agent trajectories directly \cite{jin2025search, wu2025hiprag, zhao2025beyond, javafrugalrag}. Search-R1 \cite{jin2025search} shows that purely outcome-based RL can train LLMs to interleave reasoning with multiple search queries. HiPRAG \cite{wu2025hiprag} extends this line with hierarchical process rewards that provide finer-grained credit assignment over search/non-search decisions to reduce over-search and under-search. R1-Searcher \cite{song2025r1} also incentivizes search usage via outcome-based RL in a two-stage framework, focusing on improving the model’s ability to invoke external search when needed. $\beta$-GRPO \cite{wu2025search} introduces a confidence-thresholded RL objective to mitigate suboptimal search and improve decision quality. In parallel, ZeroSearch \cite{sun2025zerosearch} addresses the high cost and instability of RL by training with simulated retrieval during RL. Our proposed approach \ourmethod~belongs to this RL-for-search family but targets a complementary bottleneck: supervision over the \emph{coverage} of the decomposed queries. We explicitly focus on shaping the quality of multi-hop query decomposition during RL via coverage-based supervision.

\section{\ourmethod - Our Method}

In this section, we first discuss the RL training framework of the LLM agent. Then we provide details on our proposed coverage reward with teacher-guided training framework.  

\subsection{Preliminaries: Search-augmented LLM Agent}

Search-augmented LLM agent produces trajectories with multi-turn retrieval and reasoning for a user query. It invokes search calls when explicitly triggered by \texttt{<}search\texttt{>} and \texttt{</}search\texttt{>}. Retrieved content is enclosed within \texttt{<}information\texttt{>} and \texttt{</}information\texttt{>} tokens, while LLM reasoning steps are wrapped within \texttt{<}think\texttt{>} and \texttt{</}think\texttt{>} tokens \cite{jin2025search}. The generation process continues until the generation of a final answer within \texttt{<}answer\texttt{>} and \texttt{</}answer\texttt{>} token or reaching a predefined action budget. Training this search-augmented LLM agent is formulated as a reinforcement learning problem, where the language model serves as a policy model that interacts with the environment. Given input question $x$, policy model $\pi_{\theta}$, and generated trajectory $y$, the objective is to maximize the expected reward over generated trajectories while constraining deviations from a reference policy:
\begin{equation}
\max_{\theta}
\mathbb{E}_{x \sim \mathcal{D},y \sim \pi_\theta(\cdot \mid x; R)}
\left[r(x,y)\right]
- \beta D_{\mathrm{KL}}\left(\pi_\theta \parallel \pi_{\mathrm{ref}}\right).
\label{eqn:rl}
\end{equation}
Here, $r(x,y)$ denotes a trajectory-level reward and $\pi_{\mathrm{ref}}$ is a frozen reference model used for regularization. Optimization is applied only to the tokens generated by the policy. We optimize the policy using Proximal Policy Optimization (PPO)~\cite{schulman2017proximal}. Let $\pi_\theta$ denote the current policy and $\pi_{\theta_{\text{old}}}$ the behavior policy used to collect the trajectories. PPO maximizes a clipped surrogate objective:
\begin{equation}
\mathcal{L}(\theta) =
\mathbb{E}_{t}\left[
\min \Big(
\rho_t(\theta) A_t,
\text{clip}\big(\rho_t(\theta),\,1-\epsilon,\,1+\epsilon\big) A_t
\Big)
\right],
\end{equation}
where $\rho_t(\theta)=\frac{\pi_\theta(y_t \mid s_t)}{\pi_{\theta_{\text{old}}}(y_t \mid s_t)}$ is the probability ratio between new and old policies, $s_t$ is the state, $\epsilon$ is a clipping hyperparameter, and $A_t$ denotes the advantage estimate. This objective limits excessively large policy updates while enabling efficient optimization of trajectory-level rewards.

\begin{table*}[ht]
    \centering 
    \caption{Performance on question answering on various datasets. \ourmethod~outperforms strong baselines like Zero-Search and Search-R1 on all multi-hop QA datasets (HotpotQA, 2wiki, MuSiQue). All training approaches reported here use Qwen2.5-3B Base model. Best and second-best results are marked in bold and underline respectively. $^\dagger/^\star$ represents in-domain/out-domain datasets. EM and FR refer to exact match and format reward respectively} 
    \label{tab:main_result} 
    \resizebox{.9\linewidth}{!}{
    \begin{tabular}{l|c|ccc|ccc|c|c} 
        \toprule 
        \textbf{Method} & \textbf{Trainset} & \multicolumn{3}{c|}{\textbf{General QA}} & 
        \multicolumn{3}{c|}{\textbf{Multi-hop QA}} & \textbf{Multi-hop QA}  & \textbf{Combined QA} \\ 
        &  & \textbf{NQ$^\dagger$} & \textbf{TriviaQA$^\star$} & \textbf{PopQA$^\star$} & \textbf{HotpotQA$^\dagger$} & \textbf{2Wiki$^\star$} & \textbf{MuSiQue$^\star$} & \textbf{Average} & \textbf{Average}\\
        \midrule 
        Direct Inference & - &  10.60 & 28.80 & 10.80 & 14.90 & 24.40 & 2.00 & 13.77 & 15.25 \\
        RAG & - &  34.80 & 54.40 & 38.70 & 25.50 & 22.60 & 4.70 & 17.60 & 30.12 \\
        IRCoT & - & 11.10 & 31.20 & 20.00 & 16.40 & 17.10 & 6.70 & 13.40 & 17.08 \\
        Search-o1 & - & 16.60 & 31.00 & 8.20 & 14.80 & 22.40 & 5.20 & 14.13 & 16.37\\
        R1 & NQ, HotpotQA & 14.20 & 34.80 & 20.80 & 19.60 & 28.40 & 6.40 & 18.13 & 20.70\\
        Zero-search & NQ, HotpotQA & \underline{43.00} & \textbf{61.60} & \textbf{44.80} & 33.80 & 34.60 & 13.00 & 27.13 & 38.47\\
        Search-R1 (EM) & NQ, HotpotQA  & 40.60 & 58.70 & 43.50 & 28.40 & 27.30 & 4.90 & 20.20 & 33.90 \\
        Search-R1 (EM, FR) & NQ, HotpotQA & 42.90 & 58.48 & 44.09 & \underline{36.04} & \underline{36.23} & \underline{13.65} & \underline{28.64} & \underline{38.94}\\
        \ourmethod & NQ, HotpotQA & \textbf{43.82} & \underline{59.52} & \underline{44.39} & \textbf{36.45} & \textbf{38.12} & \textbf{16.01} & \textbf{30.19} & \textbf{39.72} \\
        \bottomrule 
    \end{tabular}
    }
    \vspace{-1mm}
\end{table*}

\begin{table}[ht]
    \centering 
    \caption{Performance comparison on QA tasks when model is trained with only multi-hop QA dataset (HotpotQA)} 
    \label{tab:main_result_multihop_only} 
    \resizebox{\linewidth}{!}{
    \begin{tabular}{l|c|ccc|c} 
        \toprule 
        \textbf{Method} & \textbf{Trainset} & \textbf{HotpotQA} & \textbf{2wiki} & \textbf{MuSiQue} & \textbf{Average}\\
        \midrule 
        Search-R1 & NQ, HotpotQA & \underline{36.04} & \underline{36.23} & \underline{13.65} & \underline{28.64} \\
        Search-R1 & HotpotQA & 33.14 & 34.72 & 13.02 & 26.96 \\
        \ourmethod & HotpotQA  & \textbf{36.33} & \textbf{38.95} & \textbf{18.57} & \textbf{31.28} \\
        \bottomrule 
    \end{tabular}
    }
    \vspace{-2mm}
\end{table}

\subsection{Coverage Reward}

Outcome-based rewards, such as exact-match and format rewards, primarily evaluate the correctness of the final answer but do not explicitly supervise the quality of intermediate search behavior. To provide structured yet lightweight guidance, we introduce teacher generated essential search queries that serve as weak supervision signals over retrieval strategies. For each input question $x$, a frozen teacher model $\pi_T$ generates a set of essential search queries, $\mathcal{Q}^{T}(x) = \{q_1^T, q_2^T, \ldots, q_K^T\}$. These teacher generated search queries are intended to approximate the latent information requirements necessary to solve the task. Rather than prescribing a fixed reasoning trajectory, they provide a high-level representation of the key sub-questions that a successful search policy should cover. During rollout, the policy model generates its own search trajectory $y$ with sequence of search queries to interact with the environment. The corresponding set of extracted search queries from the policy-generated trajectory is $\mathcal{Q}^{P}(x,y) = \{q_1^P, q_2^P, \ldots, q_M^P\}$.

To measure how effectively the policy captures the intended retrieval intents, we compute a coverage signal based on the overlap between policy-generated queries and teacher anchors. Let $\mathcal{I}(x,y) = \mathcal{Q}^{P}(x,y) \cap \mathcal{Q}^{T}(x)$ denote the set of matched queries. This set is identified utilizing LLM judge which considers semantics and granularity as the matching criteria. Intuitively, this set represents the portion of teacher-identified information needs that the policy addresses during a rollout. We can quantify the coverage of the policy trajectory using precision and recall:
\vspace{-5pt}
\begin{equation}
Precision = \frac{|\mathcal{I}|}{|\mathcal{Q}^{P}|}, \qquad
Recall = \frac{|\mathcal{I}|}{|\mathcal{Q}^{T}|}.
\vspace{-5pt}
\end{equation}

Precision reflects how closely the policy queries align with relevant retrieval intents, penalizing redundant or irrelevant searches, while recall measures whether the policy sufficiently covers the teacher-identified sub-questions. By considering both metrics jointly, we avoid encouraging trivial strategies such as issuing many loosely related queries or producing only a small subset of essential searches. The coverage reward is computed as the harmonic mean:
\begin{equation}
r_{\text{cov}}(x,y) =
\frac{2 * Precision * Recall}{Precision + Recall },
\label{eqn:coverage}
\end{equation}

The use of the F1 formulation balances completeness and specificity, providing a smooth trajectory-level signal that encourages the model to generate concise yet comprehensive query decompositions. Unlike dense process rewards that assign credit at every reasoning step, this reward operates at the trajectory level, preserving the flexibility of autoregressive generation while still guiding retrieval behavior. The set of matched queries $\mathcal{I}(x,y)$ is inferred by an LLM judge. The final reward combines task performance and format reward with coverage guidance:
\begin{equation}
r(x,y) =
r_{\text{ans}}(x,y)
+ r_{\text{format}}(y)
+ \lambda_{\text{cov}} r_{\text{cov}}(x,y),
\end{equation}
where $r_{\text{ans}}$ and $r_{\text{format}}$ denote the exact match reward and the format reward, respectively. The outcome-based reward encourages the policy to produce correct answers following a proper format, while the coverage reward introduces an inductive bias toward generating structured and informative search queries. We optimize Eqn. \ref{eqn:rl} using the PPO method to train the search-augmented agent.

\section{Experiments}

We evaluate \ourmethod~on multi-hop and open-domain question answering benchmarks. Our goal is to assess whether explicitly supervising search query coverage improves downstream answer accuracy, search behavior, and retrieval quality in multi-hop question-answering tasks.

\noindent
\textbf{Datasets.} We conduct experiments on widely used knowledge-intensive Question Answering (QA) datasets. Natural Questions (NQ) \cite{kwiatkowski2019natural}, TriviaQA \cite{joshi2017triviaqa}, and PopQA \cite{mallen2023not} evaluate open-domain factual reasoning. HotpotQA \cite{yang2018hotpotqa}, 2WikiMultiHopQA (2wiki) \cite{ho2020constructing}, and MuSiQue \cite{trivedi2022musique} evaluate multi-hop reasoning requiring sequential retrieval and compositional search strategies. Following prior work, models are trained on NQ and HotpotQA datasets unless otherwise mentioned, and evaluated across all datasets to measure generalization.

\noindent
\textbf{Implementation Details.} Following prior works, we use Qwen2.5-3B (Base) as the policy model across all experiments. The search environment retrieves the top-$3$ passages for each generated query and appends them to the context using special delimiters. Teacher essential search queries are generated offline using Qwen2.5-72B-Instruct model. We use 8 A100 GPUs during  training and train the policy model for 600 steps. We also use Qwen2.5-72B-Instruct model as an LLM judge to identify matched queries and compute coverage reward using Eqn. \ref{eqn:coverage}. We empirically set $\lambda_{cov}$ to 0.2. For retrieval, we use the 2018 Wikipedia dump \cite{karpukhin2020dense} as the knowledge source and E5 \cite{wang2022text} as the retriever.

\begin{figure*}[ht]
    \centering
    \begin{subfigure}[c]{0.48\textwidth}
        \centering
        \includegraphics[width=\textwidth]{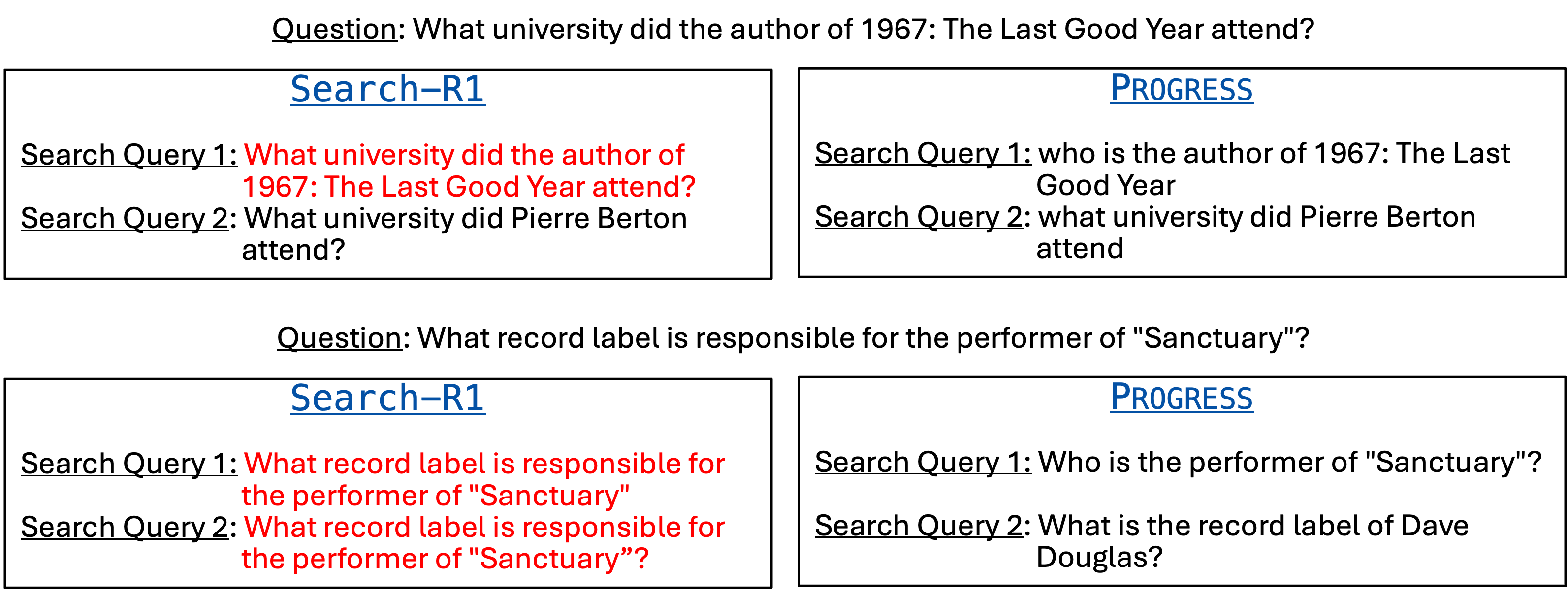}
        \caption{Two example cases where \ourmethod~successfully generates granular search query for a given complex user query compared to Search-R1. Search-R1 tends to use the complex query directly for search.}
        \label{fig:granularity}
    \end{subfigure}
    \hfill
    \begin{subfigure}[c]{0.48\textwidth}
        \centering
        \includegraphics[width=\textwidth]{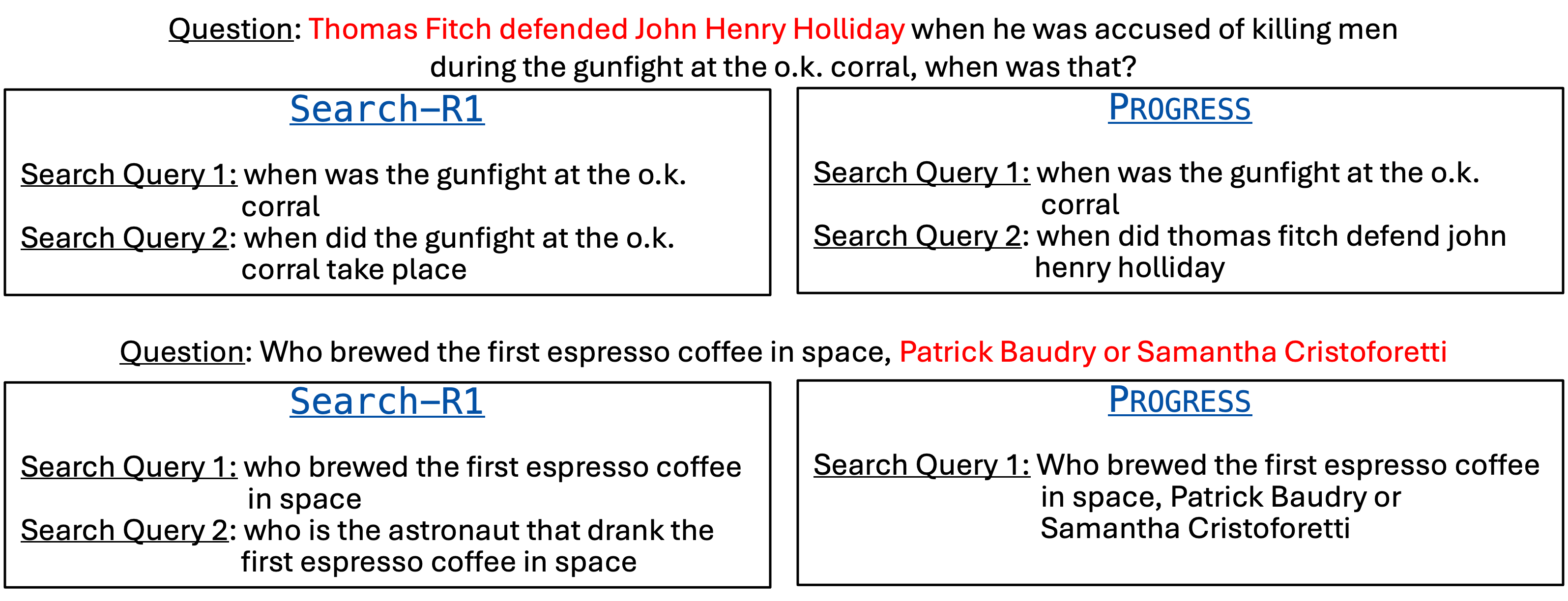}
        \caption{Two example cases where \ourmethod~successfully generates search queries that comprehensively covers all information needs of a given complex user query compared to Search-R1.}
        \label{fig:completeness}
    \end{subfigure}
    \vspace{-1mm}
    \caption{Qualitative Examples where \ourmethod~generates search query with better granularity and completeness.}
    \label{fig:qualitative}
    \vspace{-2mm}
\end{figure*}

\begin{table}[t]
    \centering 
    \caption{Search query quality analysis} 
    \label{tab:query_quality} 
    \vspace{-1mm}
    \resizebox{.8\linewidth}{!}{
    \begin{tabular}{l|c|cc} 
        \toprule 
        \textbf{Dataset} & \textbf{Method}  & \textbf{Completeness} & \textbf{Granularity} \\
        \midrule 
        \multirow{2}{*}{2wiki} & Search-R1  & 29.77 & 60.51    \\
        & \ourmethod & \textbf{40.63} & \textbf{62.04}   \\
        \midrule
        \multirow{2}{*}{MuSiQue} & Search-R1 & 18.09 & 62.06    \\        
        & \ourmethod & \textbf{24.88} & \textbf{64.38}  \\        
        \bottomrule 
    \end{tabular}
    }
    \vspace{-2mm}
\end{table}

\noindent
\textbf{Main Results.} Table \ref{tab:main_result} reports QA performance of different approaches in six datasets. Exact Match (EM) is used as the evaluation metric. We compare \ourmethod~against strong baselines that includes: i) Direct Inference: LLM answers directly without any retrieval mechanism, ii) RAG: Conventional RAG setup where retrieval is performed once based on the query, iii) Prompt-based Agentic RAG: LLM agent is prompted to achieve multi-step reasoning and search (IRCoT, Search-o1), iv) RL trained models: LLM is trained with Reinforcement Learning to enable search augmented behavior (R1, Zero-search, Search-R1). \ourmethod~consistently outperforms both training based (R1, Zero-search, Search-R1) and non-training based (RAG, IRCoT, Direct Inference) approaches for multi-hop QA datasets while retaining competitive performance for general QA datasets. The overall improvement is also evident from the averaged score from all the datasets. It has been observed that in many cases General QA dataset questions can be answered by utilizing internal knowledge. Since Zero-search approach relies on utilizing LLM internal knowledge, it has performed well on General QA datasets. To further test the significance of the coverage reward, we have trained the policy model with only multi-hop QA samples (HotpotQA) instead of commonly used combination of general QA and multi-hop QA samples. As observed in Table \ref{tab:main_result_multihop_only}, training with only multi-hop QA dataset further boost the agent's performance on multi-hop QA tasks. We gain $~5\%$ EM accuracy improvement on MuSiQue dataset compared to second best performance of Search-R1. This implies that \ourmethod~helps learn better decomposition and search strategy to boost multi-hop QA performance.

\noindent
\textbf{Search Query Quality Analysis.} The motivation behind utilizing coverage reward is to boost the comprehensiveness and the granularity of the policy generated search queries. To understand the impact of coverage reward on the policy model, we systematically analyze the quality of the search queries. We utilize 2wiki and MuSiQue dataset for this analysis. We have randomly sampled 1000 examples from each dataset, inferred the reasoning trajectories for each sample using the policy model, and extracted the search queries generated in those trajectories. To evaluate the quality of these extracted search queries, we consider the following two dimensions: i) Completeness: The decomposed search queries together should cover every key piece of information involved in the complex query. ii) Granularity: Each decomposed search query should reflect one atomic information. It should not bundle multiple atomic search queries or operations in a decomposed search query. We utilize a set of LLM judges together (GPT-4.1, GPT-5.1, and GPT-5.2) to evaluate the search query quality. Completeness is computed for each complex query; each LLM judge takes all the extracted search queries and the complex query to generate a boolean score. Then we use the mean of boolean score for each sample. For Granularity, each LLM judge generates a boolean score for each extracted search query. Then we take micro average across all search queries for the 1000 examples of the dataset. Table \ref{tab:query_quality} shows the comparison between Search-R1 and \ourmethod~for search query quality analysis. For both 2wiki and MuSiQue dataset, we observe significant improvement in Completeness and Granularity dimension. This indicates that the model trained with coverage reward results in search query with better granularity and comprehensive coverage of the complex query information need.

\begin{table}[t]
    \centering 
    \caption{Performance comparison for retrieval accuracy.} 
    \vspace{-2mm}
    \label{tab:retrieval} 
    \resizebox{\linewidth}{!}{
    \begin{tabular}{c|c|c|c|c|c} 
        \toprule 
        \textbf{Method} & \textbf{Trainset} & \textbf{HotpotQA} & \textbf{2wiki} & \textbf{MuSiQue} & \textbf{Average}\\
        \midrule 
        Search-R1 & NQ + HotpotQA & 49.37 & 37.66 & 28.94 & 38.66\\
        \ourmethod & NQ + HotpotQA & \textbf{55.54} & \textbf{43.96} & \textbf{35.38} & \textbf{44.96} \\
        \midrule
        Search-R1 & HotpotQA & 46.33 & 37.17 & 25.45 & 36.31\\
        \ourmethod & HotpotQA & \textbf{55.73} & \textbf{44.31} & \textbf{37.40} & \textbf{45.81}\\
        \bottomrule 
    \end{tabular}
    }
    \vspace{-4mm}
\end{table}

\noindent
\textbf{Retrieval Accuracy.} Success of the search-augmented agent largely depends on the retrieval performance. Correct retrieval during the reasoning process increases the chance of correct completion of the query. So beyond exact match evaluation of the answer, we also want to evaluate if coverage reward is helping to boost the retrieval performance. To evaluate retrieval performance, we extract all the retrieved passages from the trajectory and check if the golden answer string is present in any one of those retrieved passages. Table \ref{tab:retrieval} shows the percentage of queries where we have correct retrieval according to the above approach. We observe that \ourmethod~significantly improves the accuracy of retrieval in all three multi-hop QA datasets. It also shows significant improvement for training scenario where we utilize only multi-hop QA dataset to train the policy model. It suggests that guiding query decomposition using coverage reward helps improve the models' ability to craft effective and useful search queries for the downstream search mechanism.

\noindent
\textbf{Qualitative results.} Fig. \ref{fig:granularity} shows two example cases where \ourmethod~generates more granular search queries compared to Search-R1. Search-R1 tends to utilize the complex query directly for initial search, which can impact the retrieval performance negatively. Fig. \ref{fig:completeness} shows two example cases where \ourmethod~successfully generates search queries that comprehensively covers all information needs of a given complex user query compared to Search-R1. As shown in Fig. \ref{fig:completeness}, there are failure cases for \ourmethod~where it uses the complex query directly to search too.

\noindent
\textbf{Model Generalizability.} To validate the generalizability of \ourmethod, we also train a different base model (Qwen2.5-1.5B) using our framework. Table \ref{tab:main_result_qwem1.5b} shows that Qwen2.5-1.5B model trained with our framework can result in better performance on multi-hop QA task across all three datasets.



\begin{table}[t!]
    \centering 
    \caption{Performance comparison for QA task on multi-hop QA datasets using Qwen2.5-1.5B base model.} 
    \label{tab:main_result_qwem1.5b} 
    \vspace{-2mm}
    \resizebox{\linewidth}{!}{
    \begin{tabular}{l|c|ccc|c} 
        \toprule 
        \textbf{Method} & \textbf{Trainset} & \textbf{HotpotQA} & \textbf{2wiki} & \textbf{MuSiQue} & \textbf{Average}\\
        \midrule 
        Search-R1 & NQ, HotpotQA & 26.09 & 24.48 & 5.67 & 18.75 \\
        \ourmethod & NQ, HotpotQA  & \textbf{27.37} & \textbf{25.51} & \textbf{5.87} & \textbf{19.58} \\
        \bottomrule 
    \end{tabular}
    }
    \vspace{-4mm}
\end{table}

\section{Conclusion}

In this work, we introduce a coverage-guided reinforcement learning framework for training search-augmented language models. While prior approaches primarily rely on outcome-level rewards, we show that explicitly supervising the quality of the search query can provide a simple yet effective signal for improving search behavior. By leveraging teacher-generated essential search queries and measuring their overlap with policy-generated search queries, \ourmethod~encourages more comprehensive and focused retrieval strategies without requiring dense process-level supervision. Empirical results demonstrate consistent improvements in question-answering performance, query quality, and retrieval effectiveness across multiple multi-hop benchmarks. These findings highlight the importance of shaping intermediate search behavior when optimizing agentic language models. 

\newpage
\section{Generative AI Use Disclosure}
Generative AI tools like large language models have been used to assist in the editing and polishing of the text. These tools have been employed to enhance the readability, clarity, and language quality of the manuscript, without altering the core research findings, methodologies, or conclusions. All content presented in this paper has been reviewed and revised by the authors to ensure accuracy and alignment with research objectives.

\bibliographystyle{IEEEtran}
\bibliography{mybib}

\end{document}